\documentclass[a4paper,10pt]{article}

\usepackage[english]{babel} 
\usepackage[utf8]{inputenc} 

\usepackage[T1]{fontenc}
\usepackage{float}
\usepackage{graphics}
\usepackage{graphicx}
\usepackage{epsfig}
\usepackage{indentfirst}
\usepackage{amsmath, amsfonts, amssymb, amsthm, mathtools}
\usepackage{url}
\usepackage{csquotes}

\usepackage[backend=biber, style=trad-abbrv, natbib=true]{biblatex}
\newcommand{\vetx}{\boldsymbol{x}}
\newcommand{\vety}{\boldsymbol{y}}
\newcommand{\vetz}{\boldsymbol{z}}

\newcommand{\imI}{\mathbf{I}}
\newcommand{\imJ}{\mathbf{J}}

\newcommand{\vetv}{\boldsymbol{v}}
\newcommand{\vetw}{\boldsymbol{w}}

\newcommand{\bpm}{\begin{bmatrix}}
\newcommand{\epm}{\end{bmatrix}}

\usepackage{xcolor}

\title{Shortest Length Total Orders Do Not Minimize Irregularity in Vector-Valued Mathematical Morphology\footnote{This work was supported in part by the National Council for Scientific and Technological Development (CNPq) under grant no 315820/2021-7 and the S\~ao Paulo Research Foundation (FAPESP) under grant no 2022/01831-2.}}

\author{Samuel Francisco\footnote{Instituto Federal de Educação, Ciência e Tecnologia de São Paulo (IFSP), Campus Itaquaquecetuba, Itaquaquecetuba -- SP. Email: samuelfsc@ifsp.edu.br} and Marcos Eduardo Valle\footnote{Instituto de Matemática, Estatística e Computação Científica (IMECC), Universidade Estadual de Campinas (Unicamp), Campinas -- SP, Brazil. Email: valle@ime.unicamp.br}
}

\begin{document}

\maketitle

\begin{abstract}
Mathematical morphology is a theory concerned with non-linear operators for image processing and analysis. The underlying framework for mathematical morphology is a partially ordered set with well-defined supremum and infimum operations. Because vectors can be ordered in many ways, finding appropriate ordering schemes is a major challenge in mathematical morphology for vector-valued images, such as color and hyperspectral images. In this context, the irregularity issue plays a key role in designing effective morphological operators. Briefly, the irregularity follows from a disparity between the ordering scheme and a metric in the value set. Determining an ordering scheme using a metric provide reasonable approaches to vector-valued mathematical morphology. Because total orderings correspond to paths on the value space, one attempt to reduce the irregularity of morphological operators would be defining a total order based on the shortest length path. However, this paper shows that the total ordering associated with the shortest length path does not necessarily imply minimizing the irregularity. 

\noindent
{\bf Keywords}. Vector-valued mathematical morphology, irregularity issue, shortest length path.
\end{abstract}

\section{Introduction}

Mathematical morphology is a powerful mathematical theory for image processing and analysis \cite{heijmans95}. From a theoretical point of view, mathematical morphology can be very well-defined in a mathematical structure called complete lattice \cite{ronse90}. A complete lattice is a partially ordered set in which any subset has supremum and infimum. The complete lattice framework allowed the development of morphological operators for vector-valued images, including color and hyperspectral images \cite{velasco-forero14}. Nevertheless, vectors can be ordered in many different ways, and finding appropriate ordering schemes for a given task is one of the main challenges in vector-valued mathematical morphology. In this context,  the irregularity issue plays a crucial role in designing effective morphological operators \cite{chevallier16}. 

In essence, the irregularity issue follows because the topology induced by a complete lattice may not reproduce the natural topology of a metric space \cite{chevallier16}. Precisely, consider a vector-valued image $\mathbf{I}: \mathcal{D} \to \mathbb{V}$, where $\mathcal{D}$ denotes the image domain and $\mathbb{V} \subset \mathbb{R}^m$ is the value set. Let us assume that the value set $\mathbb{V}$, endowed with a partial order $\leq$, is a complete lattice. Also, let us suppose that $\mathbb{V}$ is equipped with a metric $d:\mathbb{V}\times \mathbb{V} \to [0,+\infty)$. Because morphological operators only take into account the partial order $\leq$, they can introduce irregularities if the image $\mathbf{I}$ have pixel values $\vetx,\vety,\vetz \in \mathbb{V}$ such that $\vetx \leq \vety \leq \vetz$ but $d(\vetx,\vetz) \leq d(\vetx,\vety)$. Accordingly, despite $\vetx$ is closer to $\vetz$ than to $\vety$, the inequalities $\vetx \leq \vety \leq \vetz$ suggest $\vetw$ is farther from $\vetx$ than $\vety$ \cite{Valle2022Irregularity}. Because morphological operators are defined using extrema operators, they do not consider the metric of $\mathbb{V}$, resulting in the irregularity issue. At this point, we would like to remark that the irregularity introduced by a morphological operator can be quantitatively measured using the so-called irregularity index \cite{Valle2022Irregularity}.

Because the irregularity issue follows from a disparity between the partial order and the metric, an ordering scheme that preserves the spatial neighborhoods as far as possible should result in fewer irregularities. In this context, Chevalier and Angulo proposed a total ordering scheme adapted to a given image \cite{chevallier16}. Accordingly, they minimize a cost function that, in some sense, counts the disparity between the order and the metric. On the downside, due to the high number of different pixel values in natural vector-valued images, the cost function is usually not computable and alternative approaches must be considered. A straightforward alternative could be the total order associated with the shortest length Hamiltonian path. Precisely, a total order in the set of values of an image corresponds to a Hamiltonian path on the graph associated with the image \cite{lezoray2021}.
Moreover, the shortest-length Hamiltonian path yields a total order that considers the metric of the value set, and this ordering scheme is a strong candidate to reduce the irregularity issue. Although the shortest-length Hamiltonian path has been used for designing vector-valued morphological operators \cite{lezoray2021}, this paper shows that it does not necessarily reduce the irregularity issue.

The paper is organized as follows: Section \ref{sec:bases} provides the basic concepts of mathematical morphology. The total ordering associated with the shortest-length Hamiltonian path is addressed in Section \ref{sec:permutation}. Section \ref{sec:exp} provides computational experiments on tiny color images using the total order based on the shortest-length Hamilton path and the lexicographical RGB ordering. The paper finishes with the concluding remarks in Section \ref{sec:conclusion}.

\section{Mathematical Morphology and the Irregularity Issue}\label{sec:bases}

Mathematical morphology is concerned with non-linear operators widely used for image processing and analysis \cite{heijmans95}. In a broad sense, morphological operators consider shapes and other geometrical structures in images. Complete lattices are appropriate frameworks for developing morphological operators \cite{ronse90}. A complete lattice $\mathbb{L}$ is a partially ordered set (poset) in which all subsets have a supremum and an infimum \cite{birkhoff93}. The supremum and the infimum of $X \subseteq \mathbb{L}$ are denoted by $\bigvee X$ and $\bigwedge X$, respectively.

Throughout the paper, $\mathbb{V} \subseteq \mathbb{R}^m$, $m>1$, is a complete lattice endowed with a metric $d$. A vector-valued image $\imI$ corresponds to a mapping $\imI: D \to \mathbb{V}$, where $D$ and $\mathbb{V}$ denote the image domain and the value set, respectively. We shall assume that the domain $D$ is a subset of a non-empty finite additive abelian group $(\mathcal{E}, +)$. The set 
\begin{equation} \label{eq:V(I)}
V(\imI) = \{\imI(x):x \in D \},
\end{equation} 
is the set of all vector-values of the image $\imI$.

Dilations and erosions are two elementary operators of mathematical morphology \cite{soille99}. Given a set $S \subset \mathcal{E}$, called structuring element, the dilation and the erosion of the image $\mathbf{I}$ by $S$ are the images defined by the following equations, respectively:
\begin{equation} \label{eq:ero_dil}
\delta_S(\mathbf{I})(p) = \bigvee_{\substack{s \in S\\ p-s \in D}} \mathbf{I}(p-s) \quad \mbox{and} \quad 
\varepsilon_S(\mathbf{I})(p) = \bigwedge_{\substack{s \in S\\ p+s \in D}}\mathbf{I}(p+s), 
\quad \mbox{for all} \;\;p \in D.
\end{equation}

Many morphological operators can be derived by combining the elementary morphological operators \cite{heijmans95}. For example, the compositions of dilations and erosions, known as opening and closing, possess interesting topological properties and serve as non-linear image filters \cite{soille99}. Precisely, an opening is defined as an erosion followed by a dilation of the resulting image with the same structuring element $S$. Dually, a closing is a dilation followed by an erosion by the same structuring element $S$. In mathematical terms, the opening $\gamma_{S}$ and the closing $\phi_{S}$ of an image $\imI$ by a structuring element $S$ are defined respectively by
\begin{equation} \label{eq:opening_closing}
\gamma_{S}(\mathbf{I}) = \delta_{S}\left(\varepsilon_{S}(\imI) \right) \quad \mbox{and} \quad 
\phi_{S}(\mathbf{I}) = \varepsilon_{S}\left(\delta_{S}(\imI) \right).
\end{equation}

Note that dilations and erosions are defined in \eqref{eq:ero_dil} using the supremum (or maximum) and infimum (minimum) operations. Therefore, a partial order with well-defined extrema operations is all we need to develop morphological operators. However, there are many different ways to order vectors. For example, consider a complete lattice $\mathbb{V} \subseteq \mathbb{R}^m$. The marginal ordering, also called point-wise or Cartesian product ordering, is defined as follows for vectors $\vetx=(x_1,x_2,\ldots,x_m) \in \mathbb{V}$ and $\vety=(y_1,\ldots,y_m) \in \mathbb{V}$:
\begin{equation}
    \label{eq:marginal}
    \vetx \leq_M \vety \quad \Longleftrightarrow \quad x_1 \leq_{\mathbb{R}} y_1, x_2 \leq_{\mathbb{R}} y_2, \ldots, \; \mbox{and}\; x_m \leq_{\mathbb{R}} y_m,
\end{equation}
where $\leq_{\mathbb{R}}$ denotes the usual ordering in $\mathbb{R}$. 
The supremum and infimum operations are computed component-wise  using marginal ordering. Precisely, given $\boldsymbol{X} \subseteq \mathbb{V}$, then
\begin{equation}
    \label{eq:marginal_sup_inf}
    \bigvee \boldsymbol{X} = (\bigvee X_1,\ldots,\bigvee X_m) \quad \mbox{and} \quad \bigwedge \boldsymbol{X}= (\bigwedge X_1,\ldots,\bigwedge X_m).
\end{equation}
Despite its simplicity, the marginal ordering given by \eqref{eq:marginal} often leads to false colors, or more generally, false values \cite{serra09}. For example, consider the RGB color space $\mathcal{C}_{RGB} = [0,1] \times [0,1] \times [0,1]$, where each color is composed of the red, green, and blue components. The marginal order yields
\begin{equation} \label{false_colors_example}
    \bigvee  \{(1,0,0), (0,0,1)\} = (1, 0, 1) \quad \mbox{and} \quad 
    \bigwedge \{(1,0,0), (0,0,1)\} = (0, 0, 0).
\end{equation}
In words, the supremum and the infimum of red and blue are magenta and black, respectively. Therefore, using the marginal ordering, the supremum and infimum operations may result in a color that does not belong to the set, resulting in the so-called false color. 

Total orderings can avoid the false-color problem. A partial order $\leq$ is a total order if, for any $\vetx, \vety \in \mathbb{V}$, either $\vetx \leq \vety$ or $\vety \leq \vetx$ hold true. If the value set $\mathbb{V} \subseteq \mathbb{R}^m$ endowed with a total order is a complete lattice, then the supremum and the infimum of a finite set are elements of the set. Consequently, total orders circumvent the false-color problem. The lexicographical ordering, defined as follows, is an example of a total order for $\mathbb{V} \subseteq \mathbb{R}^m$. Given $\vetx = (x_1, x_2, \ldots, x_m) \in \mathbb{V}$ and $\vety = (y_1, y_2, \ldots,y_m) \in \mathbb{V}$, we have 
\begin{equation}\label{eq:lex}
\vetx \leq_{L} \vety \iff \exists k \in \{1, 2, ..., m\}\; \text{such that}\; \forall j < k, \; x_j = y_j  \; \text{and}\; x_k \leq_{\mathbb{R}} y_k.
\end{equation}
Therefore, using the lexicographical RGB ordering, the supremum and the infimum of the red and blue colors are
\begin{equation} \label{false_colors_example2}
    \bigvee  \{(1,0,0), (0,0,1)\} = (1, 0, 0) \quad \mbox{and} \quad 
    \bigwedge \{(1,0,0), (0,0,1)\} = (0, 0, 1),
\end{equation}
which are not false colors. Actually, there are several morphological approaches based on total orders. For instance, \cite{velasco-forero14} showed some approaches based on supervised and unsupervised orders.

However, Chevallier and Angulo demonstrated that the topology induced by a total order could be inconsistent with the topology of the metric space, leading to what they referred to as the irregularity issue \cite{chevallier16}. As a consequence, the image $\imJ$ obtained by a morphological operator based on a total order may present irregularities. We would like to remark that the irregularity depends, in particular, on the total ordering. In the following section, we will pursue a total ordering that could apparently reduce the irregularity issue. 

The irregularity introduced by a morphological operator can be measured using the so-called irregularity index \cite{Valle2022Irregularity}. Briefly, the irregularity index is given by the gap between the generalized sum of pixel-wise distance and the Wasserstein metric. 
In this paper, we use the irregularity index to quantify the irregularity introduced by a morphological operator based on a total ordering \cite{Valle2022Irregularity}. 


\section{Shortest-Length Hamiltonian Path} \label{sec:permutation}

As pointed out in Section \ref{sec:bases}, the irregularity issue follows from a disparity between the total order and the metric on the set $\mathbb{V}$ of values of an image \cite{chevallier16}. In light of this remark, this section presents an approach to define a total order using the metric of $\mathbb{V}$. Namely, the total order is derived from the shortest length Hamiltonian path in the set of the pixel values of an image. 

Let $\imI: D \to \mathbb{V}$ be an image where $D=\{x_1,\ldots,x_n\}$ is a finite domain and $\mathbb{V}$ is a complete lattice. Also, let $\vetv_i = \imI(x_i)$ be the value of the image in $x_i \in D$, for $i=1,\ldots,n$. If $\leq$ is a total order on $\mathbb{V}$ then, for any $x_i, x_j \in D$, either $\imI(x_i) \leq \imI(x_j)$ or $\imI(x_j) \leq \imI(x_i)$ holds true. Hence, a total order enables the construction of an ordered list $P = [\vetv_1, \vetv_2, \ldots, \vetv_n]$ such that $\vetv_i \leq \vetv_j$ for every $i \leq j$. Moreover, the ordered list yields a Hamiltonian path in the set $V(\imI)$ given by \eqref{eq:V(I)}, that is, a path that visits each pixel value of $\imI$ exactly once.

Since the irregularity issue takes into account the metric $d$ in the value space of the image $\imI$, it is plausible to suppose that a total order in which the list $P = [\vetv_1, \ldots, \vetv_n]$ of the values of $\imI$ has the smallest Total Variation (TV) given by
\begin{equation} \label{eq:TotalVariation}
    \|P\| = \sum_{k = 2}^{n} \|\vetv_k - \vetv_{k-1} \| + \|\vetv_n - \vetv_1\|,
\end{equation}
yields more regular morphological operators.
In this context, we would like to recall that
Lézoray proposes a heuristic to obtain a total order associated with the smallest possible value of \eqref{eq:TotalVariation} \cite{lezoray2021}. 

\begin{figure}[t]
    \centering
    \begin{tabular}{rccc}
    & Image 1 & TSP order & Lex. RGB order \\
    \rotatebox{90}{$\quad$ \textbf{Dilation}} &
    \includegraphics[width=0.25\columnwidth]{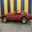} &
    \includegraphics[width=0.25\columnwidth]{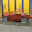} & 
    \includegraphics[width=0.25\columnwidth]{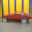} \\
     & Image 2 & $(27.89\%, 19.99)$ & $(11.07\%, 129.29)$  \\
    \rotatebox{90}{$\quad$ \textbf{Erosion}} &
    \includegraphics[width=0.25\columnwidth]{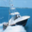} &
    \includegraphics[width=0.25\columnwidth]{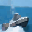} & 
    \includegraphics[width=0.25\columnwidth]{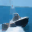} \\
    & Image 3 & $(21.59\%, 13.11)$ & $(3.42\%, 65.80)$  \\
    \rotatebox{90}{$\quad$ \textbf{Opening}} &
    \includegraphics[width=0.25\columnwidth]{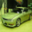} &
    \includegraphics[width=0.25\columnwidth]{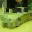} & 
     \includegraphics[width=0.25\columnwidth]{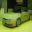} \\
    & Image 4 & $(50.28\%, 16.61)$ & $(8.23\%, 64.19)$  \\
    \rotatebox{90}{$\quad$ \textbf{Closing}} &
    \includegraphics[width=0.25\columnwidth]{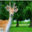} &
    \includegraphics[width=0.25\columnwidth]{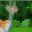} & 
    \includegraphics[width=0.25\columnwidth]{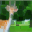} \\
    &  & $(37.21\%, 23.32)$ & $(14.83\%, 102.57)$  \\
    \end{tabular}
    \caption{ \small Dilation, erosion, opening, and closing of the tiny images 1, 2, 3, and 4, with TSP and RGB approaches, follow of the global irregularity measure $\Phi_1^g$ and the length of the Hamiltonian path.}
    \label{fig:cifar_images}
\end{figure}

Note that obtaining the smallest value of $\|P\|$ given by \eqref{eq:TotalVariation} is equivalent to finding the shortest-length Hamiltonian path or solving the traveling salesman problem (TSP). For this reason, we consider an order obtained by the shortest-length Hamiltonian path constructed by solving the TSP. Precisely, let $\imI: D \to \mathbb{V}$ be an image with $\text{Card}(D) = n$ and let $I_n = \{1, 2, \ldots, n\}$ be a set of indexes. Consider a list $P = [\vetv_1, \vetv_2, \ldots, \vetv_n]$ obtained by the shortest-length Hamiltonian path constructed in the set of image values $V(\imI)$ with the metric $d$ of $\mathbb{V}$. Note, however, that an Hamiltonian path $P$ is cyclic and we can start at any index $i$ from $1$ to $n$ in the list $P = [\vetv_1, \vetv_2, \ldots, \vetv_n]$. Because we are interested in a smoothing path, we consider a list that starts and ends in values $\vetv_1$ and $\vetv_{n}$ such that $d(\vetv_1, \vetv_n)  \geq d(\vetv_i, \vetv_{i+1})$ for every $i = 1,\ldots, n-1$. Furthermore, we take the first element such that $\|v_1\| \leq \|v_n\|$.
Thus, for every $i \in I_n$, there exists a unique $x \in D$ such that $\imI(x) = \vetv_i$, and the total order obtained is such that $\vetv_i \leq \vetv_j$ whenever $i \leq j$. From now on, we shall refer to the resulting total order as the TSP order.




\section{Computational Experiments}\label{sec:exp}

Let us investigate the irregularity introduced by morphological operators defined using the TSP order defined in Section \ref{sec:permutation}. To this end, we consider color images of size $32 \times 32$ from the CIFAR10 dataset \cite{cifar10}. We apply the dilation, erosion, opening, and closing operators defined in \eqref{eq:ero_dil} and \eqref{eq:opening_closing} with the TSP order. From a practical perspective, we calculated the TSP order using the nearest neighbor and the farthest insertion heuristics and adopted the best one. For comparison purposes, we also apply the same morphological operators defined using the lexicographical RGB order given by \eqref{eq:lex}. We quantify the irregularity introduced by the morphological operators using the irregularity index $\Phi_{1}^{g}$ between the input and output images \cite{Valle2022Irregularity}. Furthermore, we calculate the length of the Hamiltonian path associated with both the TSP and lexicographical RGB orders.

\begin{figure}[t]
    \centering
    \begin{tabular}{cccc}
   a) Path TSP of Image 1 & b) Path RGB of Image 1 \\    
    \includegraphics[width=0.5\columnwidth]{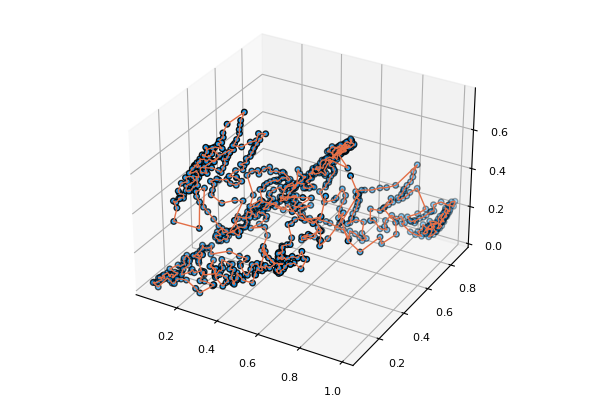} &
    \includegraphics[width=0.5\columnwidth]{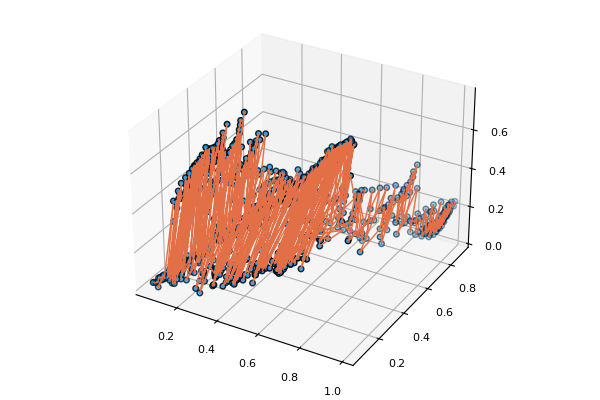}\\
    c) Path TSP of Image 3 & d) Path RGB of Image 3 \\  
    \includegraphics[width=0.5\columnwidth]{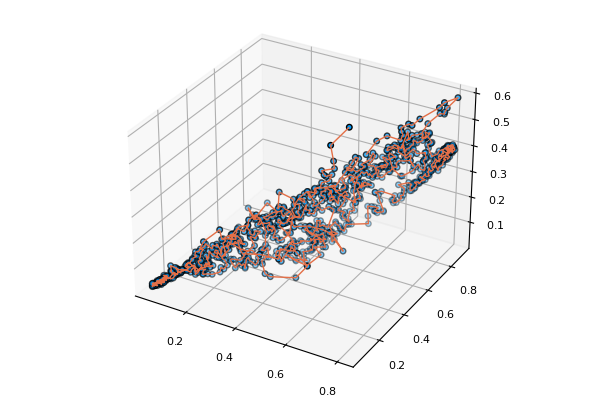} &
    \includegraphics[width=0.5\columnwidth]{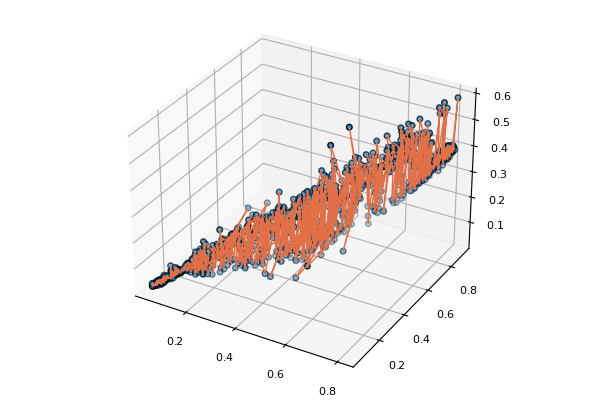}\\
     \end{tabular}
    \caption{\small Distribution of the input data obtained by the $V(\imI)$, and the respective Hamiltonian paths based on the TSP and lexicographical orderings.}
    \label{fig:Path_TSP_LEX}
\end{figure}


Figure \ref{fig:cifar_images} shows the results obtained from four images. Here, the columns show the original images, those obtained by the TSP order, and those obtained by the lexicographical RGB order, respectively. The irregularity index and the length of the Hamiltonian path are included below the operated image. Note that the irregularity for dilation, erosion, opening, and closing in the TSP approach is $27.89\%$, $21.59\%$, $50.28\%$, and $37.21\%$, respectively. The irregularity for dilation, erosion, opening, and closing in the lexicographical approach is $11.07\%$, $3.42\%$, $8.23\%$, and $14.83\%$, respectively. Thus, the images obtained by the morphological operators based on the lexicographical order are all less irregular than the image obtained using the TSP order. However, the  Hamiltonian paths for the TSP order are all shorter than those obtained from the lexicographical order. Therefore, minimizing the path does not imply minimizing the irregularity.

Figure \ref{fig:Path_TSP_LEX} shows, in the RGB cube, the values of the set $V(\imI)$ and the Hamiltonian paths obtained from images 1 and 3 using TSP and lexicographical orders. Note that the lexicographical order indeed prioritizes the first component in constructing the path. Note from Figure \ref{fig:Path_TSP_LEX} that the Hamiltonian paths obtained from the TSP order are indeed shorter than those obtained from the lexicographical RGB order. 

\section{Concluding Remarks}\label{sec:conclusion}

The irregularity issue in mathematical morphology on vector-valued images follows because the information contained in a total order is too weak to reproduce the topology induced by a metric in the space of values \cite{chevallier16}. In light of this important result, one natural attempt would be deriving a total order from a metric. Indeed, Chevallier and Angulo introduced a cost function to measure the quality of a total order with respect to a metric \cite{chevallier16}. Despite its theoretical significance, minimizing such cost function is a computationally intractable problem for natural vector images. Although not addressing the irregularity issue, we note that Lezoray also proposed a total order using a metric in the value set  \cite{lezoray2021}. Precisely, because a total order gives a Hamiltonian path in a graph derived from a vector-valued image, Lezoray defines a total order based on the shortest-length Hamiltonian path.



In this paper, we addressed the question: Does a total order associated with the shortest-length Hamiltonian path yield more regular  morphological operators? The answer to this question is no. Accordingly, the computational experiments reported in Section \ref{sec:exp} compares morphological approaches based on a total order derived from the shortest-length Hamiltonian path and the lexicographical RGB. The total order associated with the shortest-length Hamiltonian path is obtained by solving the traveling salesman (TSP) problem using either the nearest neighbor or the farthest insertion heuristics and referred to the TSP order. Although the TSP order yields Hamiltonian paths shorter than those associated with the lexicographical order, the morphological operators based on the TSP order are more irregular than those obtained using the lexicographical order. Figures \ref{fig:cifar_images} and \ref{fig:Path_TSP_LEX} confirm the negative answer to the question.

For future works, one can pursue a total order that minimizes the irregularity index. For example, given an image $\imI$ and a morphological operator $\psi$, one can use an evolutionary algorithm to obtain a total order which minimizes the irregularity of the image $\imJ = \psi(\imI)$.

\printbibliography

\end{document}